\documentclass[sigconf]{acmart}

\copyrightyear{2019}
\acmYear{2019} 
\setcopyright{iw3c2w3}
\acmConference[WWW '19]{Proceedings of the 2019 World Wide Web Conference}{May 13--17, 2019}{San Francisco, CA, USA}
\acmBooktitle{Proceedings of the 2019 World Wide Web Conference (WWW '19), May 13--17, 2019, San Francisco, CA, USA}
\acmPrice{}
\acmDOI{10.1145/3308558.3314135}
\acmISBN{978-1-4503-6674-8/19/05}

\usepackage{booktabs} 
\usepackage{graphicx}
\usepackage{ifthen}
\usepackage{multirow}
\definecolor{forestgreen}{HTML}{009B55}
\definecolor{sepia}{HTML}{671800}
\definecolor{midnightblue}{HTML}{006795}
\definecolor{orangered}{HTML}{ED135A}

\usepackage{balance}

\begin{document}

\title{Automated Fact Checking in the News Room}

\author{Sebasti\~{a}o Miranda}
\affiliation{%
  \institution{Priberam Labs}
  \streetaddress{Alameda D. Afonso Henriques, 41, 2º, 1000-123, Lisbon, Portugal}
  \city{Lisbon}
  \country{Portugal}
  \postcode{1000-123}
}
\email{ssm@priberam.com}

\author{David Nogueira}
\affiliation{%
  \institution{Priberam Labs}
  \streetaddress{Alameda D. Afonso Henriques, 41, 2º, 1000-123, Lisbon, Portugal}
  \city{Lisbon}
  \country{Portugal}
  \postcode{1000-123}
}  
\email{dan@priberam.com}

\author{Afonso Mendes}
\affiliation{%
  \institution{Priberam Inform\'{a}tica, S.A.}
  \streetaddress{Alameda D. Afonso Henriques, 41, 2º, 1000-123, Lisbon, Portugal}
  \city{Lisbon}
  \country{Portugal}
  \postcode{1000-123}
}
\email{amm@priberam.com}

\author{Andreas Vlachos}
\affiliation{%
  \institution{Dep. of Computer Science and Technology, University of Cambridge}
  \city{Cambridge}
  \country{United Kingdom}
}
\email{andreas.vlachos@cst.cam.ac.uk}

\author{Andrew Secker}
\affiliation{%
  \institution{BBC News Labs}
  \streetaddress{BBC Broadcasting House, Portland Place, London, W1A 1AA}
  \city{London}
  \country{United Kingdom}
}
\email{andrew.secker@bbc.co.uk}

\author{Rebecca Garrett}
\affiliation{%
  \institution{BBC Monitoring}
  \streetaddress{BBC Broadcasting House, Portland Place, London, W1A 1AA}
  \city{London}
  \country{United Kingdom}
}
\email{becky.garrett@bbc.co.uk}

\author{Jeff Mitchel}
\affiliation{%
  \institution{Psychology Dep., University of Bristol}
  \city{Bristol}
  \country{United Kingdom}
}
\email{jeff.mitchell@bristol.ac.uk}

\author{Zita Marinho}
\affiliation{%
 \institution{Priberam Labs}
  \streetaddress{Alameda D. Afonso Henriques, 41, 2º, 1000-123, Lisbon, Portugal}
  \postcode{1000-123}
}
\affiliation{%
 \institution{Institute of Systems and Robotics, IST}
  \streetaddress{Alameda D. Afonso Henriques, 41, 2º, 1000-123, Lisbon, Portugal}
  \city{Lisbon}
  \country{Portugal}
  \postcode{1000-123}
}
\email{zam@priberam.com}

\renewcommand{\shortauthors}{Miranda et al.}

\begin{abstract}
Fact checking is an essential task in journalism; its importance has been highlighted due to recently increased concerns and efforts in combating misinformation. In this paper, we present an automated fact checking platform which given a claim, it retrieves relevant textual evidence from a document collection, predicts whether each piece of evidence supports or refutes the claim, and returns a final verdict. We describe the architecture of the system and the user interface, focusing on the choices made to improve its user friendliness and transparency. 
We conduct a user study of the fact-checking platform in a journalistic setting: we integrated it with a collection of news articles and provide an evaluation of the platform using feedback from journalists in their workflow.
We found that the predictions of our platform were correct 58\% of the time, and 59\% of the returned evidence was relevant.
\end{abstract}

%
%

\ccsdesc[500]{Computing Methodologies~AI}
\ccsdesc[300]{Computing Methodologies~NLP}
\ccsdesc[100]{Computing Methodologies~Information Extraction}

\ccsdesc[500]{Human Centered Computing~HCI}
\ccsdesc[300]{Human Centered Computing~HCI design and evaluation methodologies}
\ccsdesc[100]{Human Centered Computing~User studies}
\ccsdesc[100]{Human Centered Computing~Usability testing}

\ccsdesc[500]{Information Systems~Information Retrieval}
\ccsdesc[300]{Information Systems~Evaluation of retrieval results}
\ccsdesc[100]{Information Systems~Presentation of retrieval results}

\ccsdesc[300]{Information Systems~Retrieval tasks and goals}
\ccsdesc[100]{Information Systems~Question answering}
\keywords{Fact Checking; Computational Journalism; Media Tools} 

\maketitle

\section{Introduction}
\label{sec:intro}

Grounding information on reliable sources is a daunting experience, given the increasing amount of information circulating the web and other media platforms. Nevertheless, checking the veracity of claims is crucial for preserving trust in news sources.
Manual verification of claims is a tedious task, that consumes a lot of time and effort from journalists and professional fact-checkers, and it typically requires searching for specific entities and content over large amounts of unstructured text~\citep{Cohen2011,Hassan2015}.

The rising interest in fact checking has led to the development of a number of approaches and tools automating the task or parts of it, with the motivation of facilitating the work of journalists, and interested readers more broadly~\cite{Graves2018AFC,thorne2018review}.
The most popular effort, ClaimBuster~\cite{Hassan2017}, proposed a fact checking platform which detects factual claims that are worth checking and then uses  APIs to query search engines and databases (Google and Wolfram Alpha respectively). It also compares claims against previously fact checked ones in its database. The latter approach is also used by the system developed by Full Fact, Live, which is used to fact check repeated or paraphrased claims.\footnote{\url{https://fullfact.org/blog/2017/jun/automated-fact-checking-full-fact/}} 
Neither of these approaches is able to fact check previously unchecked claims, while the queries through existing commercial APIs are not tailored to fact checking, thus retrieving information that is not necessarily relevant.
Other approaches rely instead on the detection of rumours based on the spread of readership over social media~\citep{Qazvinian2011}. However, a rumorous claim is not necessarily false, and vice versa \citep{Zubiaga:2018:DRR:3186333.3161603}.
There is also work that checks claims against tables, such as those released by the World Bank\footnote{\url{https://data.worldbank.org/}}, using trained classifiers to select the appropriate tuple \cite{Thorne2017}. However such approaches are typically restricted to fact checking  numerical claims against tabular sources.
Finally, some approaches aim at providing a pipeline of tools for information retrieval~\citep{Wiedemann}, but do not go as far as to provide an actual fact check mechanism.

In this work, we propose an automated fact checking platform that checks claims by identifying sentences providing evidence in a large document collection. These sentences are classified as supporting, refuting or only related to the claim, and then combined into a final verdict using a state-of-the-art neural network-based approach~\citep{W18-5515}. The model is trained on the recently released FEVER dataset \cite{Thorne2018}, a large scale fact checking dataset derived from Wikipedia comprising 185K claims. Unlike previous work, our model is able to check novel claims without relying on a database of fact checks. Also, the evidence retrieved and classified as supporting or refuting provides a justification of the verdict, which is likely to be relevant not only to assess the overall correctness of the platform, but also as part of the fact checking research conducted by journalists.
Our retrieval model considers more information about a specific claim than generic search engines by leveraging  information from words and named entities present in the dataset to obtain the best matching evidence for each claim (see Figure~\ref{fig:example_claim_ent}). 
\begin{figure}
    \centering
    \begin{tabular}{ p{8.3cm} }
    \textbf{Claim:}\\ \textit{\textcolor{purple}{\textbf{Tesla}} builds car factory in \textcolor{blue}{\textbf{Shanghai}}. }\\ \hline
    \vspace{0.01cm}
    \textbf{Evidence:}\\ Electric carmaker \textcolor{purple}{\textbf{Tesla}} has signed an agreement with Chinese authorities to build a factory in \textcolor{blue}{\textbf{Shanghai}}. ``We hope it will be completed very soon,'' \textcolor{purple}{\textbf{Tesla}} chief \textcolor{olive}{\textbf{Elon Musk}} said.\\
    \end{tabular}
    \caption{Example of claim and evidence. Extracted named entities in bold: organizations (purple), locations (blue) and people (green).}
    \label{fig:example_claim_ent}
    \vspace{-0.2in}
\end{figure}

We thoroughly evaluate the platform through user testing by journalists from the British Broadcasting Corporation (BBC) in the context of their workflow. 
Of the total 488 evidence passages retrieved by the system, the journalists reported that 58\% were relevant and 59\% were accurately classified as {\sc supports}/{\sc refutes}/{\sc other}.

 
Our platform can be applied to document collections beyond the ones used in this paper and our findings should help inform future research in automated fact checking and computational journalism more broadly.


\section{Fact Checking System}
\begin{figure}[!t]
 \centering
    \includegraphics[page=1,width=0.85\columnwidth, trim={0.4in 0.65in 4.5in 0.1in}, scale=0.48,clip]{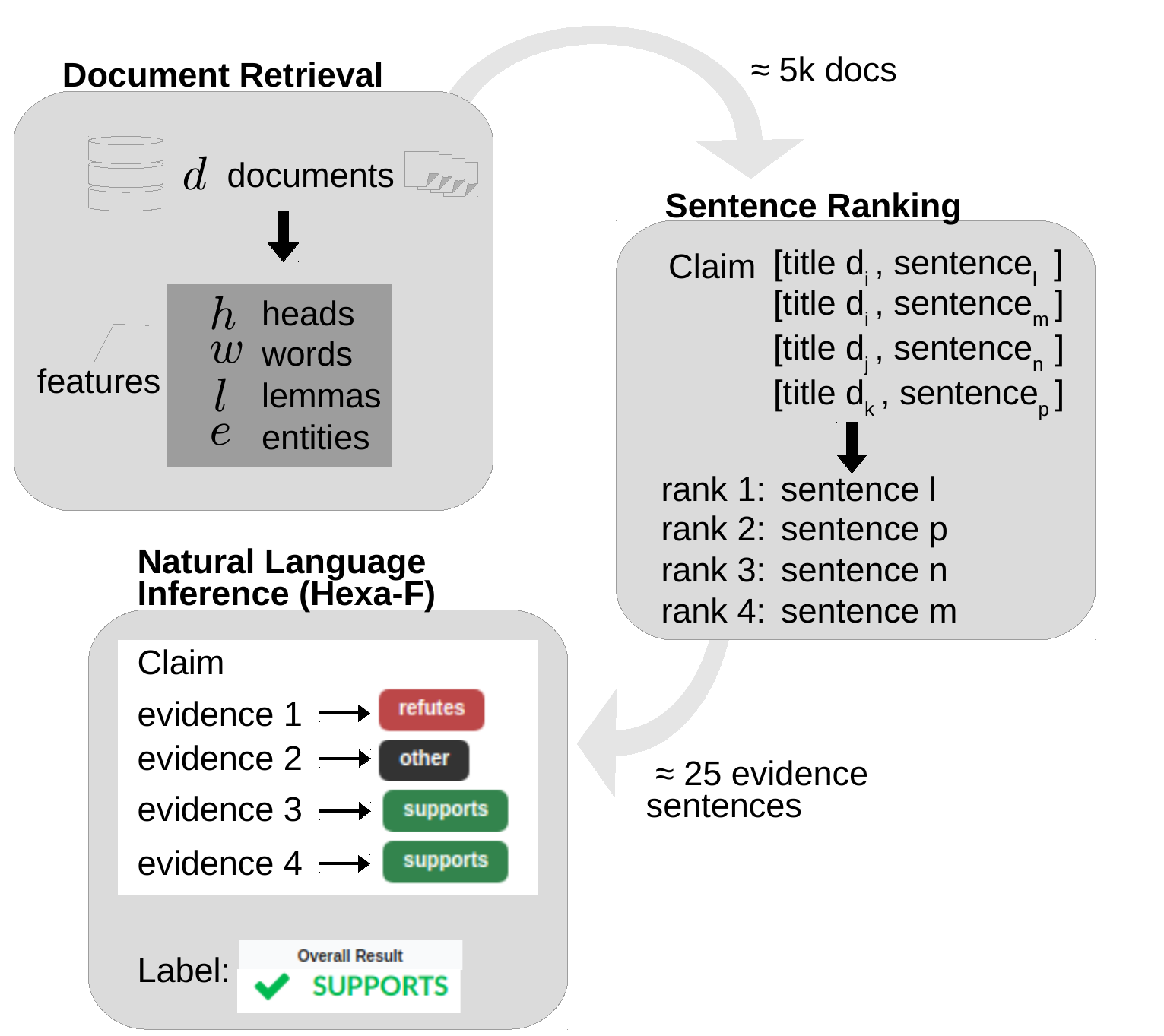}
   \caption{Fact-checking model comprised of three components: a) Document retrieval (top left), b) Sentence ranking (middle right), c) NLI prediction model (bottom).}
  \label{fig:system}
  \vspace{-0.15in}
\end{figure}

Our fact-checking system comprises three main components: a document retrieval step, a sentence ranking, and a classification model, as shown in Figure~\ref{fig:system}.

Initially we retrieve documents from a collection of news articles, using a customized search engine based on inverse indexing and retrieval using the Okapi-BM25 algorithm~\citep{RobertsonWB00}. Data is incrementally indexed from word and entity-level inverted indexes.
The document retrieval component searches for documents whose features best match the claim. 
In the end of this step, we end up with approximately $10K$ documents related to the claim. 

Following this, we rank the sentences in these documents according to predefined token and feature matching rules. We compute the cosine similarity of the claim with each best candidate sentence, using word embeddings trained on the One Billion Word Benchmark corpus~\citep{billionbenchmark}, and select those above a threshold tuned during development.  
 This step aims at providing only sentences with high relevance to the claim, reducing the number of potential evidence that may be only marginally related with the claim.

In the final component, we classify each of the extracted candidate evidence in terms of whether they support, refute or are just related to the claim (other). We employ the  natural language inference (NLI) model from the Hexa-F system \cite{W18-5515} (one of the best performing systems in the FEVER shared task \citep{W18-5500}) to classify the relation between the selected evidence sentences and the claim, one of {\sc supports}/{\sc refutes}/{\sc other}, and a similar label which expresses whether the combined set of evidence sentences {\sc supports}, {\sc refutes} or is simply related to the claim.

\subsection{Engineering Considerations}
\label{sec:eng}
\begin{figure*}[!t]
 \centering
    \includegraphics[page=2, width=\textwidth, trim={0.2in 0.7in 1.9in 0.8in}, scale=0.9]{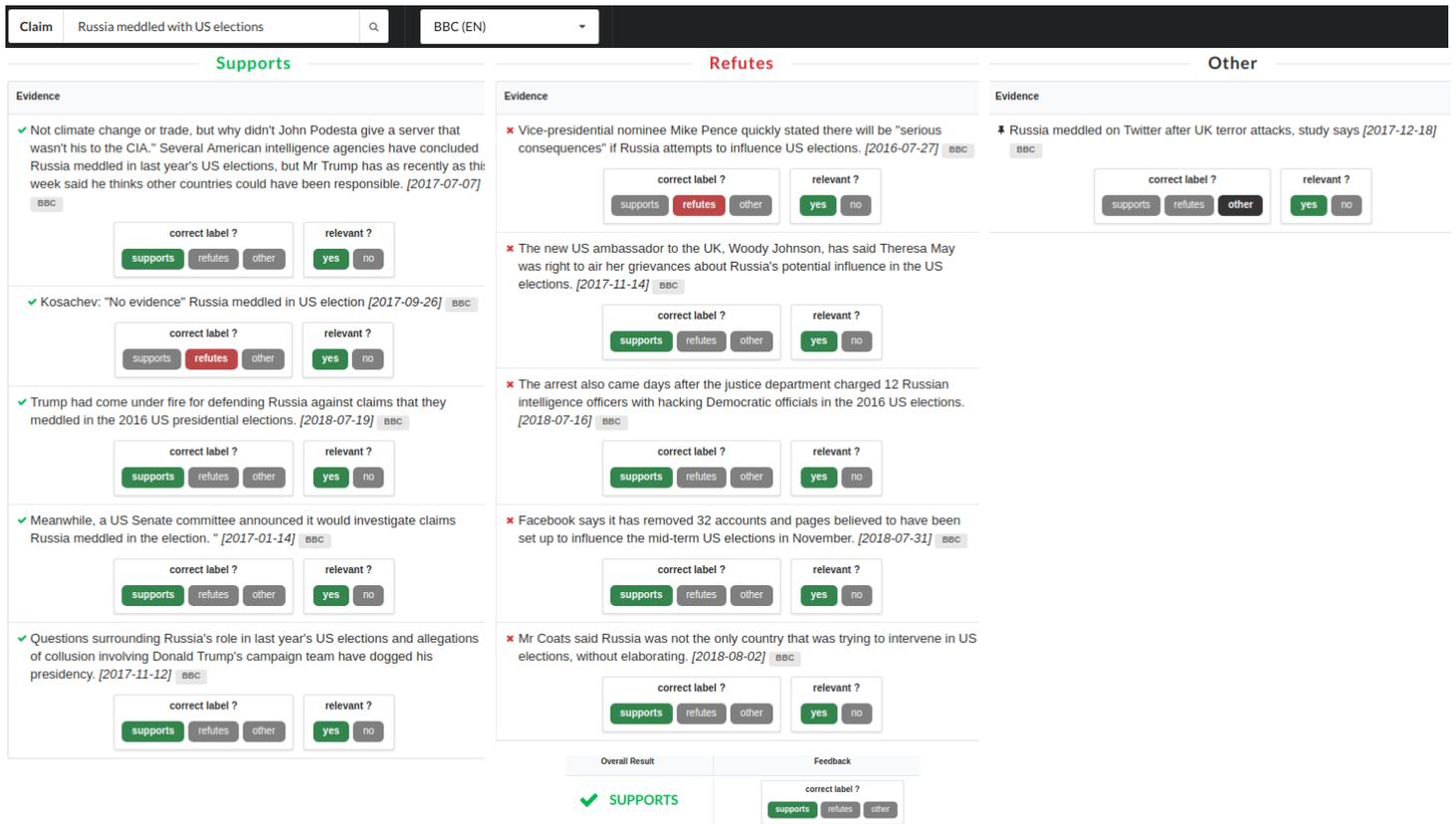}
   \caption{Example of the fact checking interface. Claim: ``Russia meddled with US elections'' (top). Five maximum evidence sentences for each column: {\sc supports}/{\sc refutes}/{\sc other} (middle). Example of final system decision and feedback buttons (bottom).}
  \label{fig:demopic}
  \vspace{-0.1in}
\end{figure*}

In the first document retrieval component, we use inverted indexes to extract relevant documents, in Figure~\ref{fig:demopic}. We select all documents that match features in the claim, such as lemmas, words and extracted named entities~\citep{summaedl}. 
This document retrieval step takes about 50 ms to retrieve around 5k documents out of a dataset with about $445k$ documents.

In the second component, we rank all the sentences from the 5k documents based on how well each sentence feature $\phi(s_i)_j$ matches the claim $\phi(c)_j$, using a positional ranking approach.
We use a ranking score based on ordered distances between $N$ matching features in the sentence $i$: $S_1(s_i,c) =\sum_{j=1}^N\exp{\left( -d_{i,j}\right)}$, where $d_{i,j}=\textrm{pos}(\phi(s_i)_j)-\textrm{pos}(\phi(s_i)_{j-1})$ represents the word distance between the position ($\textrm{pos}$) of each $j$-th feature.
This score is maximized if all words in the sentence match the claim exactly; it decreases exponentially with the word distance between matches. This part takes on average 336 ms. 
Next, we filter the sentences based on the following rules: we keep only those sentences with length $<500$ words; those that contain all the entities mentioned in the claim, as well as 
novel words that were not previously seen in previously selected sentences (less than 90\% overlap with all words previously encountered). The novelty filter increases the diversity in the 
evidence passed to the entailment step.

In the end of the second component, we re-rank the sentences by averaging
the feature matching score of each sentence $S_1$ 
and 
the cosine similarity between the claim and the sentence embeddings $S_2=\textrm{cos}(s_i,c)$. We obtain these embeddings via a weighted average of all words in the sentence, weighted by term-frequency/inverse document frequency.
To improve the performance of the sentence re-ranking component, we added the title of the document to each sentence and considered the weighted sum of all words in the title and sentence combined. This part takes about 652 ms, although it could be easily parallelizable. On average, 76 sentences are retrived per claim after these steps.

We further removed all sentences with an averaged similarity score below a given threshold ($\left(S_1+S_2\right)/2$<$0.6$), to ensure high quality evidence. In the end, about 25 sentences on average are selected.
After re-ranking, all selected sentences are used as input in the NLI model (third component) which takes about 738 ms to predict labels for each of the sentences and the overall label for the claim.

  \vspace{-0.025in}
\section{User Interface}
\label{sec:ui}
The interface allows end-users to input a claim (black bar at the top of Figure~\ref{fig:demopic}), and receive a set of evidence sentences as output. The evidence is displayed in three columns: the top five ranking sentences that are in favour of the claim on the left, the top five that are against the claim in the middle, and the top five other sentences related to the claim on the right. 
In the bottom, a final overall label is presented to the user as either other, supporting or refuting the claim (label at the bottom).
The interface allows users to directly evaluate it, providing feedback for evidence sentence w.r.t.\ the correctness of the label (``correct label?'') and its relevance (``relevant?''), as well as the correctness of the overall prediction.
We provide a video with an example of a user interacting with the platform via the interface in \url{https://vimeo.com/309336679}.

\vspace{-0.025in}
\section{User evaluation design}
\label{sec:exp}


11 BBC journalists provided feedback on the overall system and on the classification model. The journalists were asked to interact with the system by providing factual claims and evaluating the output of the model.
For each claim, up to 15 evidence sentences were presented to them, 5 per category, each classified as {\sc supports}, {\sc refutes} or {\sc other}.
For each sentence, the journalists provided feedback on two aspects: relevance and correctness.

To assess correctness, for each evidence sentence returned by the system, the journalist inputs the label which, according to his/her research on the subject, would be the correct one via the buttons in the ``correct label'' box (see Figure~\ref{fig:demopic}). This measures primarily the accuracy of the entailment component of the system, assuming that the sentences returned are all related.
For the final classification of the model (see Figure~\ref{fig:demopic} (bottom)) they also assessed the overall prediction, to whether the claim was globally supported or refuted considering all the retrieved evidence.

To assess relevance, the journalists were also asked to provide feedback as to whether they found each sentence returned relevant (see the buttons in the ``relevant ?'' box in Figure~\ref{fig:demopic}).
This part of the feedback aims to evaluate the quality of the retrieved evidence: whether it helps the journalists fact-check the input claim, regardless of the classification label attributed by the system. It also serves as a proxy to measure the precision of the retrieval component, as all sentences shown to the journalists should be relevant (ideally).

\begin{figure}[t]
 \centering
    \includegraphics[page=7, trim={0.in, 1.8in, 2in, 0.in},scale=0.7,clip]{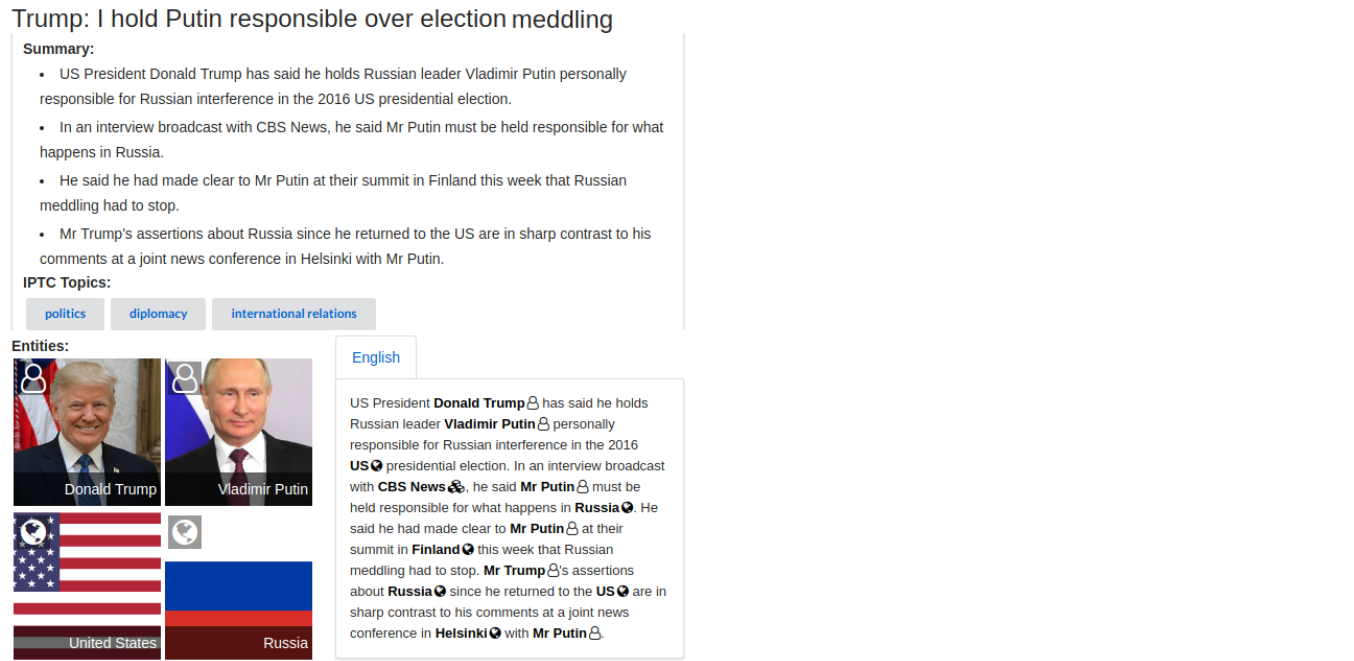} 
   \caption{Example of additional information for each evidence. Original document (right) with entities in bold.}
  \label{fig:demo_ent}
  \vspace{-0.25in}
\end{figure}

The journalists also had access to (i) a Question Answering (QA) \citep{weissenborn2018jack} tool that could serve as an additional source of information, (ii) the full document's text with annotated entities \citep{summaedl}.
The QA implemented in the platform was used by the journalists, but was not evaluated in this paper.
We provide an example of the additional information, available for each extracted evidence sentence in Figure~\ref{fig:demo_ent}.

\section{Results}
\label{sec:results}

Table~\ref{tab:jeval} summarizes the results of the system is the user evaluation conducted. We show  precision
for each class (row:Relevant) measuring the proportion of the retrieved evidence sentences that were deemed relevant by the journalists,
and precision for the predictions of the platform for the relation of each evidence sentence to the claim (row:Evidence) and the global prediction for the claim taking all the evidence into account 
(Overall result presented in Figure~\ref{fig:demopic}) (row:Global).
The platform was evaluated on 67 claims in total, with a total of 488 evidence sentences retrieved: 30\% supporting, 14\% refuting and 56\% related to the claim, as classified by our platform.

\begin{table}[t]
\centering
 \center{
\begin{tabular}{|l|cccc|}
    \hline
    \multirow{2}{*}{Precision} & \multicolumn{4}{c|}{Class} \\ 
     & \textsc{supports} & \textsc{refutes} & \textsc{other} & \textsc{all} \\ \hline
  Relevant & 71\%& 69\%& 49\% & 59 \%\\    
  Evidence Correctness & 48\%& 27\%& 70\% & 58 \%\\ 
  Global Correctness& 56\%& 26\%& 31\% & 42 \%\\
  \hline
\end{tabular}
}
  \caption{ User evaluation on the full dataset.\label{tab:jeval}}
  \vspace{-0.4in}
\end{table}

In 71\% of the claims checked by the journalists, it was reported that the evidence shown in the \textsc{supports} column was relevant, and so was for 69\% of the evidence in the \textsc{refutes} columns. We consider these results to be very encouraging, given the difficulty of the task. Retrieving evidence that contradicts a given claim, is not usually as simple as retrieving related evidence by feature matching. Introducing different information, e.g.\ entities, dates, actions, etc., that could refute the claim requires more complex language understanding methods.
We further observe that the precision in predicting evidence as {\sc supports} is 48\% in the full dataset and increases to 67\% in the subset of evidence deemed relevant (not shown in the table). The same trend is observed for the {\sc refutes} label from 27\% to 39\%. This result suggests that the retrieval component still requires some improvement, especially for retrieving evidence refuting the claim. Strategies beyond 
feature matching are needed to improve the retrieval of relevant but opposing arguments.
The precision of the classifier predicting the global label of the claim given the evidence also requires further improvement.

Additionally, we received textual feedback from the journalists about the 
overall quality of the platform. They mentioned that it is helpful for fact checking, despite not being always accurate, both in the retrieval of relevant evidence and in their evaluation.
An interesting remark mentioned possible improvements for handling time-frames and dates. For instance, claims using the present tense should refer to current events, while those mentioned in the past tense together with dates should refer to that specific time-period only. Also they suggested that presenting the evolution of results over time would be very helpful to substantiate the claim.
Handling time constraints in the retrieval process is a very interesting and challenging research direction. 
On the whole, the journalists reported that the system has a lot of potential to help their work.  

This user testing was extremely useful both for BBC journalists to experiment with state-of-the-art technology and for us to receive feedback to improve our platform in the future. 

\vspace{-0.05in}
\section{Conclusions and Future Work}
\label{sec:conclusion}
This paper introduces a novel fact checking platform aimed to assist journalists in their investigative work-flow. Our platform can be used for search of supporting and refuting evidence regarding factual claims. We evaluated using on a journalistic corpus with testing by eleven journalists, which found it to yield relevant results in 59\% of the retrieved evidence. The performed user study provided very fruitful feedback to direct future work in automated fact checking. 
Suggested improvements such as handling temporal remarks, pose an interesting issue that we found very relevant to advance research in the field of information retrieval for fact checking.



\begin{acks}
The authors would like to thank James Thorne for insightful discussion and the BBC journalists for preforming the user testing and for their valuable feedback. 
This work is supported by the EU H2020 SUMMA project (grant agreement \textnumero~688139), and by Lisbon Regional Operational Programme (Lisboa 2020), under the Portugal 2020 Partnership Agreement, through the European Regional Development Fund (ERDF), within project INSIGHT (\textnumero~033869).

\end{acks}

\bibliographystyle{ACM-Reference-Format}
\balance
\bibliography{factChecking}

\end{document}